# Mini bot 3D: A ROS based Gazebo Simulation

B. Udugama – M00734040

***Abstract*— The recent adoption of the Robot Operating System (ROS) as a software standard in robotics has contributed to novel solutions for several problems on the area. One such problem is known as Simultaneous Localization and Mapping (SLAM) with autonomous navigation, for which a number of algorithms from different classes are available as ROS packages ready to be used on any compatible robot. Many anticipated applications of autonomous mobile robots require for them to navigate in diverse complex environments without support from exterior infrastructures. To perform this on-board navigation, the robot must make use of the available sensor technologies and fuse the most reliable data respective to the present environment in an adaptive manner and optimize the algorithm parameters prior to the actual implementation to reduce the workaround time. This paper will review recent efforts to develop onboard navigation systems which can seamlessly transition between outdoor and indoor environments and different terrains seamlessly using Gazebo simulator with ROS integration. The methodologies surveyed include SLAM, Odometry and Localisation. An overview of the state-of-the-art is provided with a focus on approaches which are adaptive to dynamic sensor uncertainty, dynamic objects and dynamic scenes. The experiences reported on this work should provide insight for roboticists seeking an Autonomous SLAM solution for indoor applications.***

## I. Introduction

RECENTLY, the Robot Operating System (ROS) [1] was introduced to facilitate the development of software for robotics. ROS has contributed to novel solutions for several problems in robotics [2] by providing tools that enable inter robot communication and hardware abstraction. One such problem is known as Simultaneous Localization and Mapping (SLAM) [3], [4], in which the robot has to compute a map of its surroundings at the same time that its localization within the map is estimated. The mass adoption of ROS as the de facto standard in robotics has originated a number of SLAM algorithms [5], [6], [7], [8], [9] ready to be used on any ROS compatible robot. Solutions for SLAM are an important step towards enabling

Robots to operate without human intervention. In general, the algorithms that solve this problem are built upon a theoretical probabilistic approach [10] which models how to best estimate the robot state given noisy measurements from one or more sensors. Because of the intrinsicity of each sensor type, SLAM can be tackled by different algorithms classes, such as SLAM for range sensors [8], [9] and SLAM for visual sensors [5], [6], [7]. This last category is further divided according to the number of cameras used into monocular SLAM [5] and stereo SLAM [6], [7]. Considering that this variety of algorithms is easily accessible as ROS packages, there is often a need to evaluate and choose a appropriate solution aiming a particular application. In light of this, our work aims to experimentally evaluate a selected number of ROS compatible SLAM algorithms and thus, to provide a preliminary guidance for beginners and experienced roboticists on the selection of a SLAM solution for indoor applications. Because of the large availability and low cost of LIDAR sensors (e.g. RP LIDAR), analysis to algorithms that can operate utilizing this type of sensor. Nevertheless, LIDAR sensors are versatile and therefore can be used without any additional effort.

Then the most important navigation stack involvement will be compared with respect to few critical parameters and in the end come up with a explanative conclusion to point out best practices with ROS and Gazebo simulations.

## II. Kinematics of Differential Drive Motor

{Xl, Yl} coordinates indicate the inertial reference frame and {Xr, Yr} coordinates indicate the robot frame. The robot position which is P {x y θ} as expressed in the cartesian coordinates of inertial frame.Fig 1.5 Representation of robots on cartesian coordinates Inertial and Robot frame relation:

Both the individual wheels contribute the robotic motion, it is presumed that the wheels do not slide. Which can be expressed as Non-Holonomic Constraint.

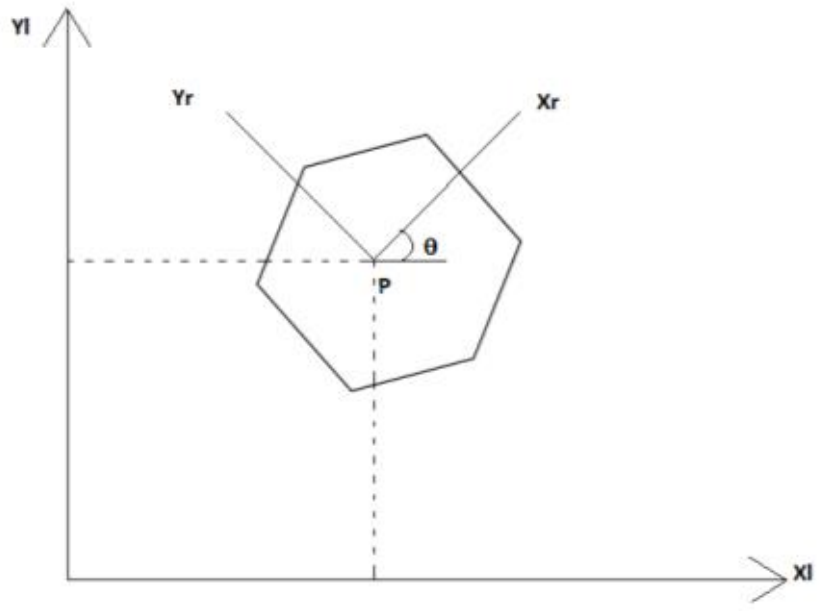

Fig 1. Mobile platform with differential drive

$$\dot{x}\sin\theta - \dot{y}\cos\theta = 0 \qquad (2)$$

$Wr$ and $Wl$ are the angular velocities of the actual robot motions with respect to the right and left wheels.

Consider the motion of the wheels against the translation speed at P in the $Xr$ direction. The one-wheel movies at a velocity of $V1 = r * Wr$ while the other wheel stays constant $V_2 = 0$. P is halfway between the two wheels so it will move with half the speed $V_x = \frac{1}{2} * r * Wr$. In a differential drive robot, these two contributions can simply be added to calculate the $Vx$. Consider, for example, a differential robot in which each wheel spins with equal speed but in opposite directions. The result is a stationary, spinning robot. As expected, $Vx$ will be zero in this case. The value of $Vy$ is even simpler to calculate. Neither wheel can contribute to sideways motion in the robot's reference frame, hence $Vy$ is always zero. Finally, we must compute the rotational component $w$ of the robot. Once again the contributions of each wheel can be computed independently and just added. Consider the right wheel (wheel 1). Forward spin of this wheel results in counter clockwise rotation at point P. Recall that if wheel 1 spins alone, the robot pivots around wheel 2. The rotation velocity $w1$ at P can be computed because the wheel is instantaneously moving along the arc of a circle of radius d. $w1 = \frac{r}{d} * Wr$. The same calculation applies to the left wheel, with the exception that forward spin results in clockwise-rotation at point P : $w2 = \frac{r}{d} * Wr$.

Mapping between Robot velocities to wheel velocities is given as follows:

$$v = v_X = v_1 + v_2 = r\left(\frac{w_R + w_L}{2}\right) ; v_Y = 0$$

$$\omega = w_1 + w_2 = r\left(\frac{w_R - w_L}{d}\right) \qquad (3)$$

Where, r = radius of wheel and d = axial distance between wheels.

## III. Creating a Robot Model : The Differential Drive Mini Bot

A differential wheeled robot will have two wheels connected on opposite sides of the robot chassis which is supported by caster wheels or dummy wheels. The wheels will control the speed of the robot by adjusting individual velocity. If the two motors are running at the same speed it will move forward or backward. If a wheel is running slower than the other, the robot will turn to the side of the lower speed. There are two supporting wheels called dummy wheels that will support the robot and freely rotate according to the movement of the main wheels. The UDRF model of this robot is present in the ROS package. The final robot model is shown as follows:

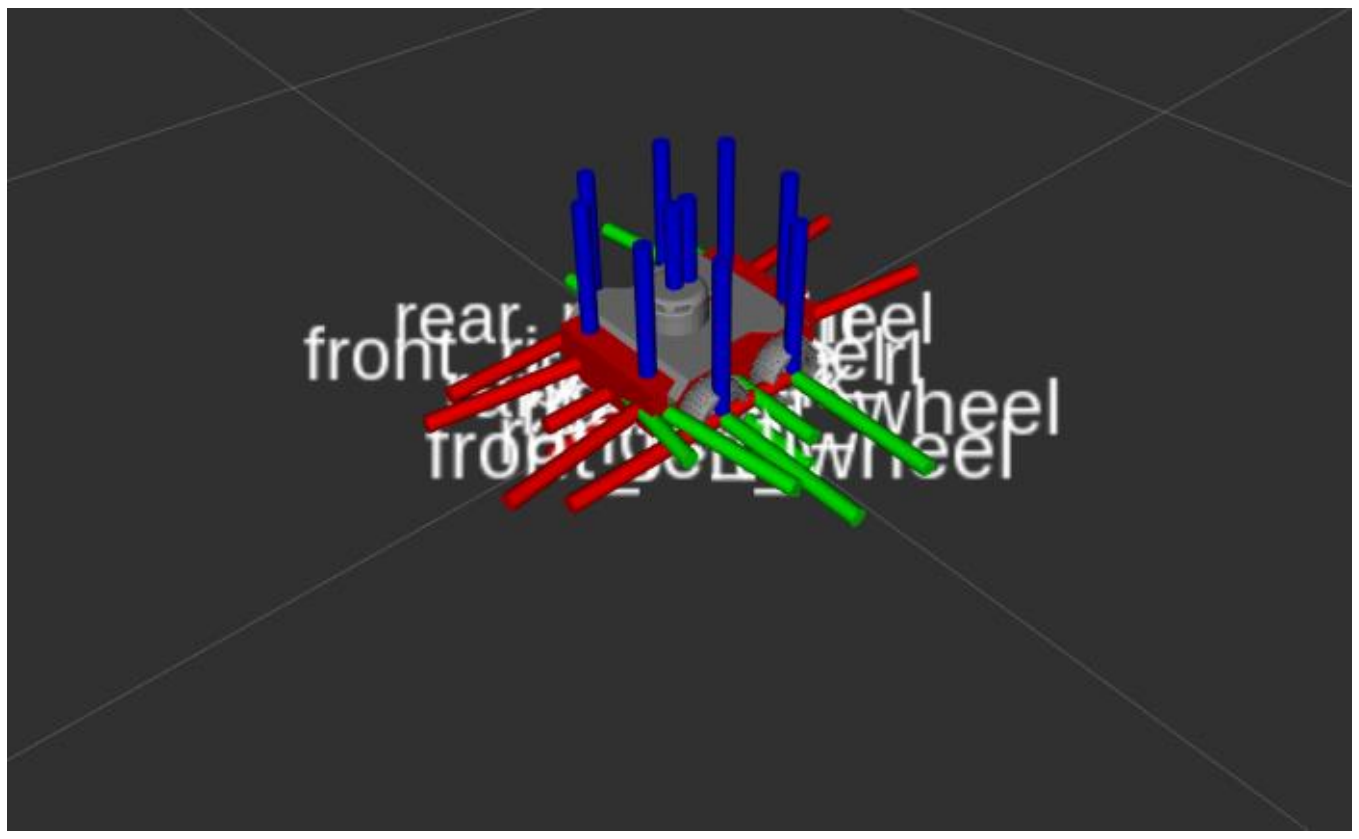

Fig 2. TF frames in robot

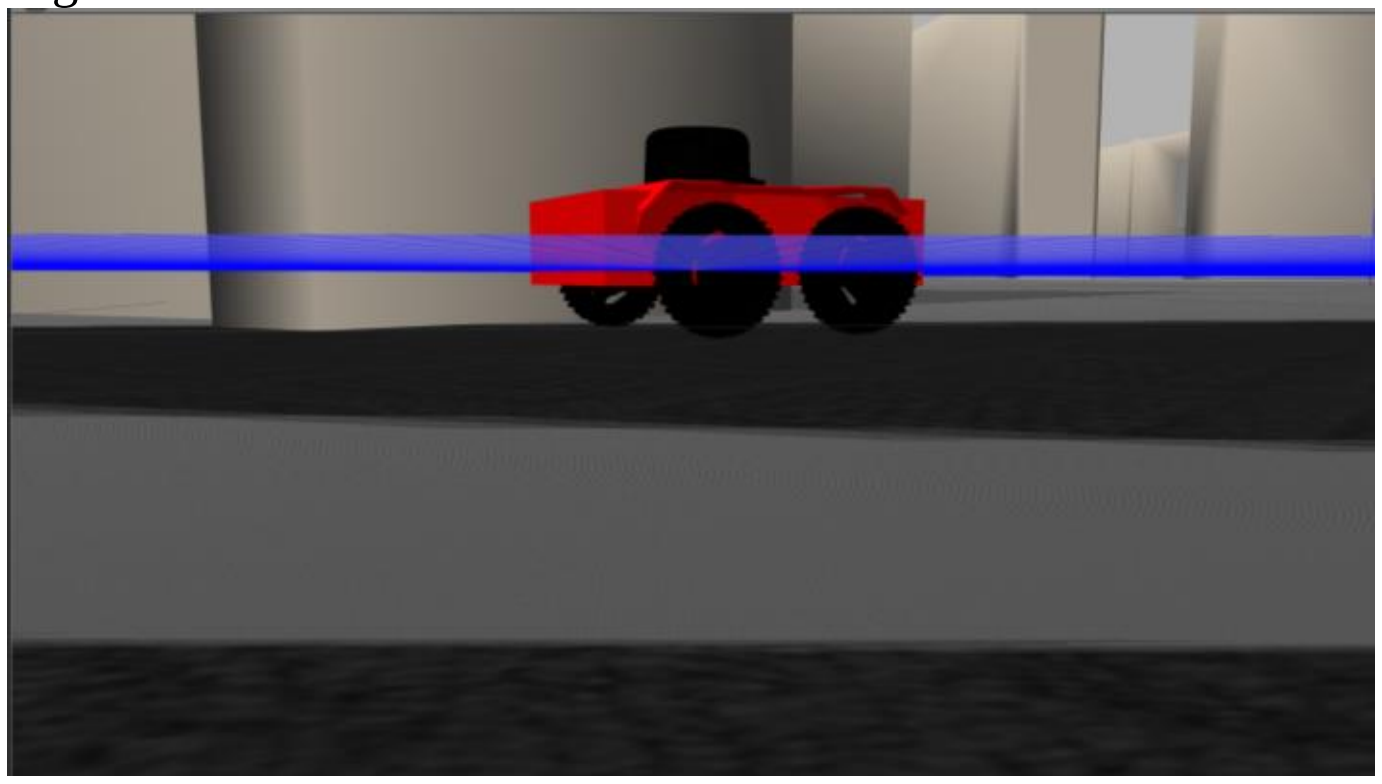

Fig 3. Gazebo robot model

The preceding robot has eleven joints and eleven links. The two main joints are four wheel joints and the other fixed things with LIDAR, and one fixed joint by base foot print. Here is the connection graph of this robot:

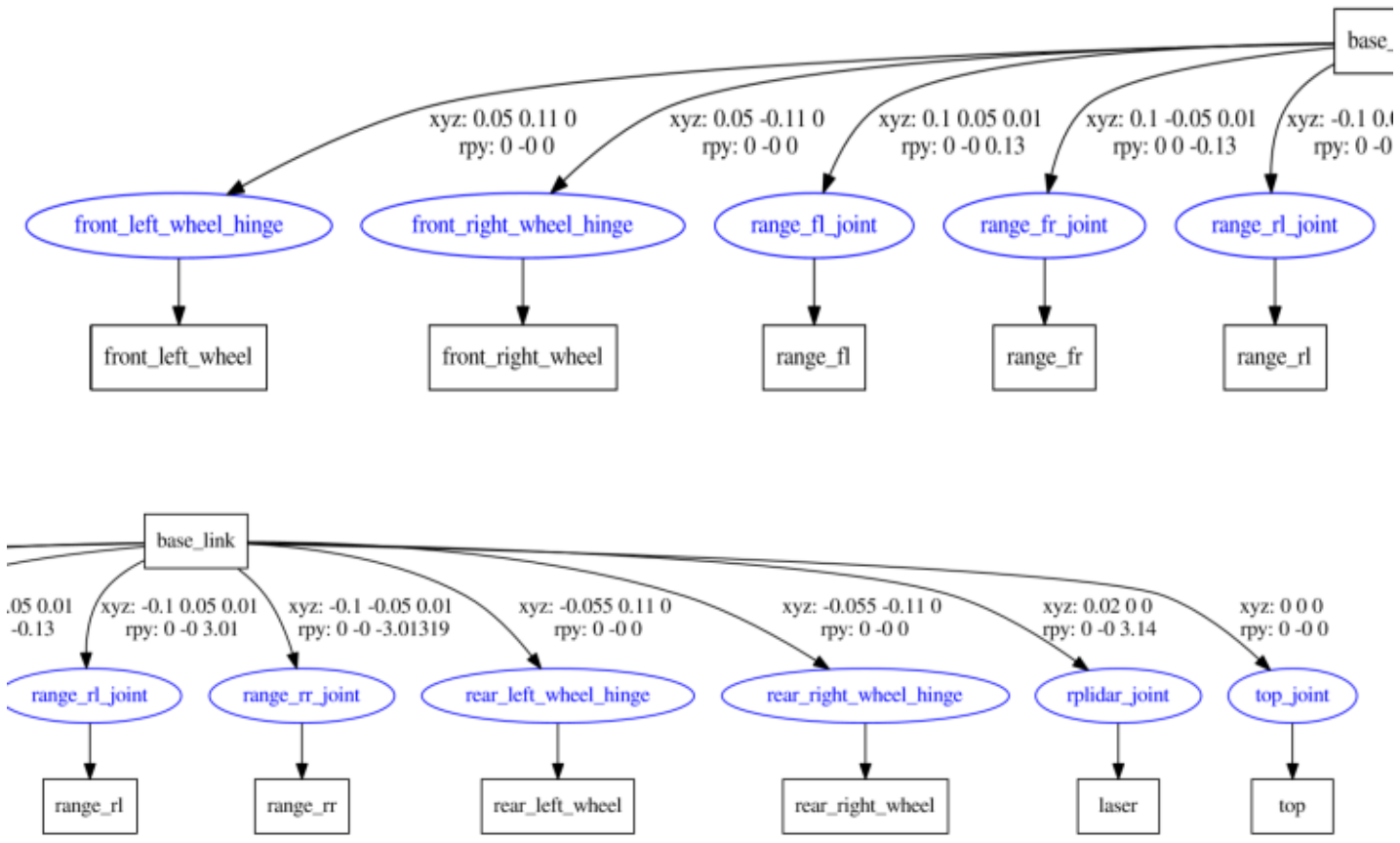

Fig 4. Parent child relation in mobile platform

The important sections of code in the UDRF file is discussed here. The UDRF file name called minibot3D.xacro is placed inside the urdf folder of the ROS package. The first section of the UDRF file is given here. This xacro file contains the definition of the wheel and its transmission; if we use this xacro file, then we can avoid writing two definitions for the two wheels. This xacro definition can be repeatedly used because wheels are identical in shape and size:

Mentioned here are the Gazebo parameters associated with a wheel. It can mention the frictional coefficient and stiffness co-efficient using the gazebo reference tag. Also need to mention the transmission tag of each wheel; the macro of the wheel is as follows:

```
<joint name="${lr}_wheel_hinge" type="continuous">
    <parent link="chassis"/>
    <child link="${lr}_wheel"/>
  <origin xyz="${-wheelPos+chassisLength/2}
${tY*wheelWidth/2+tY*chassisWidth/2} ${wheelRadius}" rpy="0 0 0" />
    <axis xyz="0 1 0" rpy="0 0 0" />
    <limit effort="10000" velocity="10000"/>
    <joint_properties damping="0.0" friction="0.0"/>
  </joint>
```

Using the preceding lines, the wheels on the left and right of the robot base from front. The robot base is cube in shape as shown in the preceding figure. The inertia calculating macro is given here. This xacro snippet will use the mass, radius, and height of the cylinder and calculate inertia using this equation:

```
  <!--macro name="cylinder_inertia" params="m r h">
   <inertia  ixx="${m*(3*r*r+h*h)/12}" ixy = "0" ixz = "0"
     iyy="${m*(3*r*r+h*h)/12}" iyz = "0"
     izz="${m*r*r/2}"
   />
  </macro>
```

The launch file definition for displaying this root model in RViz is given here. The launch file is named view_mobile_robot.launch:

```
<?xml version="1.0" encoding="UTF-8"?>
<launch>

  <rosparam command="load" file="$(find
joint_state_controller)/joint_state_controller.yaml" />
  <node name="joint_state_controller_spawner" pkg="controller_manager"
type="spawner" output="screen" args="joint_state_controller" />

  <param name="robot_description" command="$(find xacro)/xacro --
inorder '$(find minibot3d_sim)/urdf/minibot3d.xacro'"/>

  <node name="minibot3d_spawn" pkg="gazebo_ros" type="spawn_model"
output="screen" args="-urdf -param robot_description -model minibot3d" />

  <node name="robot_state_publisher" pkg="robot_state_publisher"
type="state_publisher"/>

</launch>
```

The only difference between the arm UDRF file is the change in the name; the other sections are the same. We can view the mobile robot using the following command:

***Roslaunch minibot3d_sim minibot3d_gazebo.launch***

## IV. Generating the Map

### A. LASER SCANNER

The laser scanner was added on the top of Gazebo in order to perform high-end operations such as autonomous navigation using this robot. That extra code section needed to be added in minibot3d.xacro to have the laser scanner on the robot:

```
<!-- rplidar Laser -->
  <joint name="rplidar_joint" type="fixed">
    <axis xyz="0 1 0" />
    <origin xyz="0.02 0 0" rpy="0 0 3.14"/>
    <parent link="base_link"/>
    <child link="laser"/>
  </joint>

  <link name="laser">
    <collision>
      <origin xyz="0 0 0.058" rpy="1.5707 0 4.71"/>
      <geometry>
        <mesh
filename="package://minibot3d_sim/meshes/rplidar.dae"
scale="0.001 0.001 0.001" />
      </geometry>
    </collision>

    <visual>
      <origin xyz="0 0 0.058" rpy="1.5707 0 4.71"/>
      <geometry>
        <mesh
filename="package://minibot3d_sim/meshes/rplidar.dae"
scale="0.001 0.001 0.001" />
      </geometry>
    </visual>

    <inertial>
      <mass value="1e-5" />
      <origin xyz="0 0 0.058" rpy="1.5707 0 4.71"/>
      <inertia ixx="0" ixy="0" ixz="0" iyy="0" iyz="0"
izz="0" />
    </inertial>
  </link>
```

In this section,the Gazebo ROS plugin file called libgazebo_ros_laser.so is used to simulate the laser scanner.

### B. MOVING THE MOBILE ROBOT

The robot is working with a differential robot with two wheels, and dummy wheels. The complete characteristics of the robot should model as the Gazebo-ROS plugin for the simulation. Luckily, the plugin for a basic differential drive is already implemented. In order to move the robot in Gazebo, we should add a Gazebo ROS plugin file called libgazebo_ros_control.so to get the differential drive behavior in this robot. Here is the complete code snippet of the definition of this plugin and its parameters:

```
  <gazebo>
   <plugin name="skid_steer_drive_controller"
filename="libgazebo_ros_skid_steer_drive.so">
     <robotNamespace>${namespace}/</robotNamespace>
     <updateRate>10.0</updateRate>
     <robotBaseFrame>${namespace}/base_link</robotBaseFrame>
     <wheelSeparation>0.1</wheelSeparation>
     <wheelDiameter>0.08</wheelDiameter>
     <torque>0.5</torque>
```

```
      <leftFrontJoint>front_left_wheel_hinge</leftFrontJoint>
      <rightFrontJoint>front_right_wheel_hinge</rightFrontJoint>
      <leftRearJoint>rear_left_wheel_hinge</leftRearJoint>
      <rightRearJoint>rear_right_wheel_hinge</rightRearJoint>
      <topicName>cmd_vel</topicName>
      <commandTopic>cmd_vel</commandTopic>
      <broadcastTF>true</broadcastTF>
      <odometryTopic>odom</odometryTopic>
      <odometryFrame>${namespace}/odom</odometryFrame>
      <covariance_x>0.000100</covariance_x>
      <covariance_y>0.000100</covariance_y>
      <covariance_yaw>0.010000</covariance_yaw>
    </plugin>
  </gazebo>

  <gazebo>
    <plugin name="gazebo_ros_control"
filename="libgazebo_ros_control.so">
      <legacyModeNS>true</legacyModeNS>
    </plugin>
  </gazebo>
```

## C. THE ROS TELEOP NODE

The ROS teleop node publishes the ROS Twist command by taking keyboard inputs. From this node, we can generate both linear and angular velocity and there is already a standard teleop node implementation available. Here is the launch file called keyboard_teleop.launch to start the teleop node:

```
<?xml version="1.0"?>
<launch>
<!-- differential_teleop_key already has its own built in velocity smoother -->
  <node pkg="minibot3d_sim" type="diff_wheeled_robot_key"
name="diff_wheeled_robot_key"  output="screen">

    <param name="scale_linear" value="5" type="double"/>
    <param name="scale_angular" value="5" type="double"/>
    <remap from="turtlebot_teleop_keyboard/cmd_vel" to="/cmd_vel"/>

  </node>
</launch>
```

Launch the Gazebo with complete simulation settings using the following commands:

1. roslaunch minibot3d_sim minibot3d_rviz_gm-apping.launch

Start the teleop node:

2. roslaunch minibot3d_sim minibot3d_teleop.launch

Start RViz to visualize the robot state and laser data:

3. rosrun rviz rviz

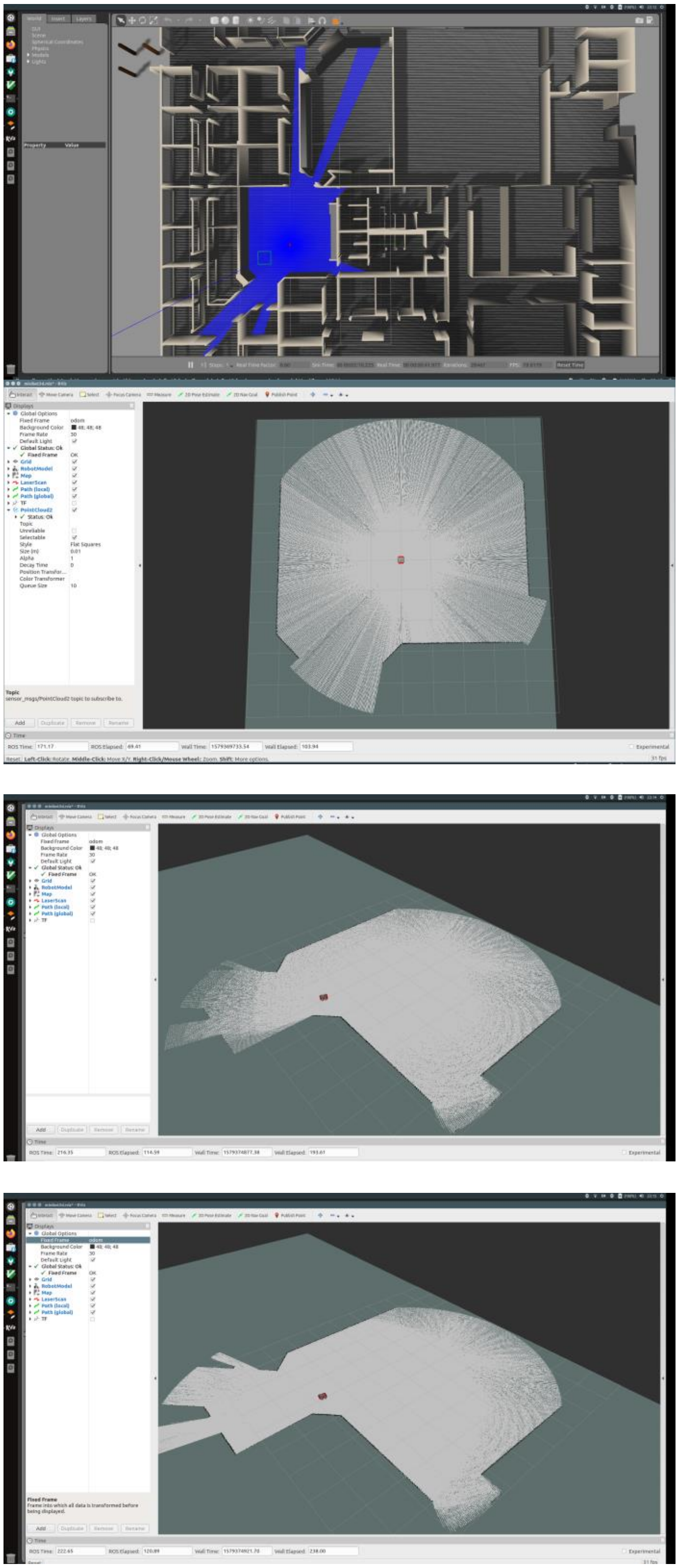

Fig 5. Map generation steps and outputs

In the teleop terminal, use some keys (U, I, O, J, K, L, M, "," , ".") for direction adjustment and other keys (Q, Z, W, X, E, C, K, space key) for speed adjustments.

We can save the built map using the following command. This command will listen to the map topic and save into the image. The map server package does this operation.

**rosrun map_server map_saver -f test_map**

Here willo is the name of the map file. The map file is stored as two files: one is the YAML file which contains the map metadata and the image name, and second is the image which has the encoded data of the occupancy grid map.
The saved encoded image of the map is shown next. If the robot gives accurate robot odometry data, we will get this kind of precise map similar to the environment. The accurate map improves the navigation accuracy through efficient path planning.

### *D. IMPLEMENTING AUTONOMOUS NAVIGATION USING AMCL AND A STATIC MAP*

The ROS AMCL package provide nodes for localizing the robot on a static map. The amcl node subscribes the laser scan data, laser scan based maps, and the tf information from the robot. The amcl node estimates the pose of the robot on the map and publishes its estimated position with respect to the map. If we create a static map from the laser scan data, the robot can autonomously navigate from any pose of the map using AMCL and the move_base nodes. The first step is to create a launch file for starting the amcl node. The amcl node is highly customizable; we can configure it with a lot of parameters.

A typical amcl launch file is given next. The AMCL node is configured inside the amcl.launch.xml file which is in the diff_wheeled_robot_gazebo/launch/ include package. The move_base node is also configured separately. The map file we created in the gmapping process is loaded here using the map_server node.

```
<?xml version="1.0"?>
<launch>
  <master auto="start"/>

  <!-- Map server -->
  <arg name="map_file" default="$(find
minibot3d_sim)/maps/test_map2.yaml"/>
  <node name="map_server" pkg="map_server" type="map_server"
args="$(arg map_file)" />

  <!-- Localization -->
  <node pkg="amcl" type="amcl" name="amcl" output="screen">
    <remap from="scan" to="/scan"/>
    <param name="odom_frame_id" value="odom"/>
    <param name="odom_model_type" value="diff-corrected"/>
    <param name="base_frame_id" value="base_link"/>
    <param name="update_min_d" value="0.1"/>
    <param name="update_min_a" value="0.2"/>
    <param name="min_particles" value="500"/>
  </node>

  <!-- Move base -->
  <node pkg="move_base" type="move_base" respawn="false"
name="move_base" output="screen">
    <rosparam file="$(find
minibot3d_sim)/config/costmap_common_params.yaml" command="load"
ns="global_costmap" />
    <rosparam file="$(find
minibot3d_sim)/config/costmap_common_params.yaml" command="load"
ns="local_costmap" />
    <rosparam file="$(find
minibot3d_sim)/config/local_costmap_params.yaml" command="load" />
    <rosparam file="$(find
minibot3d_sim)/config/global_costmap_params.yaml" command="load" />
    <rosparam file="$(find minibot3d_sim)/config/trajectory_planner.yaml"
command="load" />

    <remap from="cmd_vel" to="cmd_vel"/>
    <remap from="odom" to="odom"/>
    <remap from="scan" to="/scan"/>
    <param name="move_base/DWAPlannerROS/yaw_goal_tolerance"
value="1.0"/>
    <param name="move_base/DWAPlannerROS/xy_goal_tolerance"
value="1.0"/>

  </node>
</launch>
```

Here willo is the name of the map file. The map file is stored as two files: one is the YAML file which contains the map metadata and the image name, and second is the image which has the encoded data of the occupancy grid map.

Yaml:

```
image:
/home/bavantha/simulation_assignment/catkin_ws/src/minibot3d_sim/maps/te
st_map2.pgm
resolution: 0.010000
origin: [-5.000000, -15.560000, 0.000000]
negate: 0
occupied_thresh: 0.65
free_thresh: 0.196
```

## V. CONTROLLING ROBOT MOTION : PYTHON SCRIPT

This project proposed an integrated Robot, showing that is able to map the environment successfully and able to localize the Robot. Mini Bot is a robotic system that programmed using ROS and Gazebo simulation environment. After the map data acquirement, navigation goals are directed by a pyhton script.

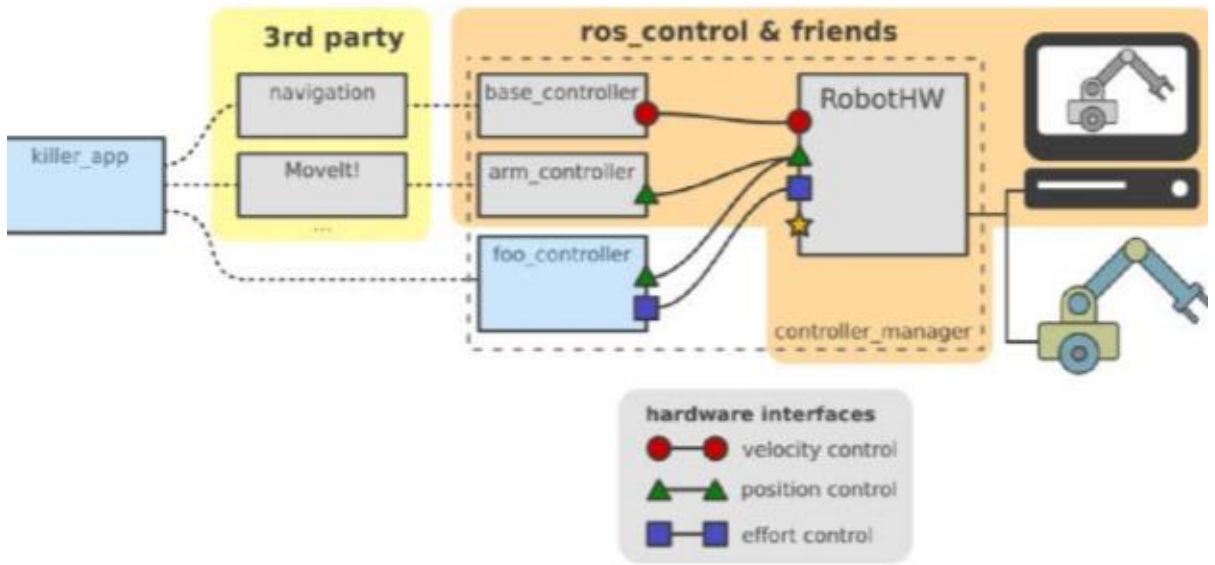

Fig 6. Ros Controller systems

The RViz 2D Nav Goal button was used to send a goal position to a robot for moving it from point A to B. Actionlib client and Python based ROS APIs are used to command the robot. Following is a sample package and node for communicating with Navigation stack move_base node. The move_base node is SimpleActionServer. Command can be sent and cancel the goals to the robot if the task takes a lot of time to complete. The following code is SimpleActionClient for the move_base node, which can send the x, y, and theta from the command line arguments.

Actionlib is imported to create the client-server communication link between ROS, Gazebo and “simple_move” node.

```
#!/usr/bin/env python
import rospy
import actionlib
```

The distance and the orientation towards goal should be inserted thorough the action server. In this particular node, it will capture the input from terminal itself and pass it to the ROS and Gazebo system.

Input captured from terminal and called the following function:

```
simple_move(sys.argv[1],sys.argv[2])
```

That function will transmit the corresponding value to the navigation stack. Only the required statements are mentions for the clarity. Important steps of the custom goal function will be :

- Initialising the node
- Declare a action client
- Just pass the goal Disatcne and Oreintation to the Navigation stack

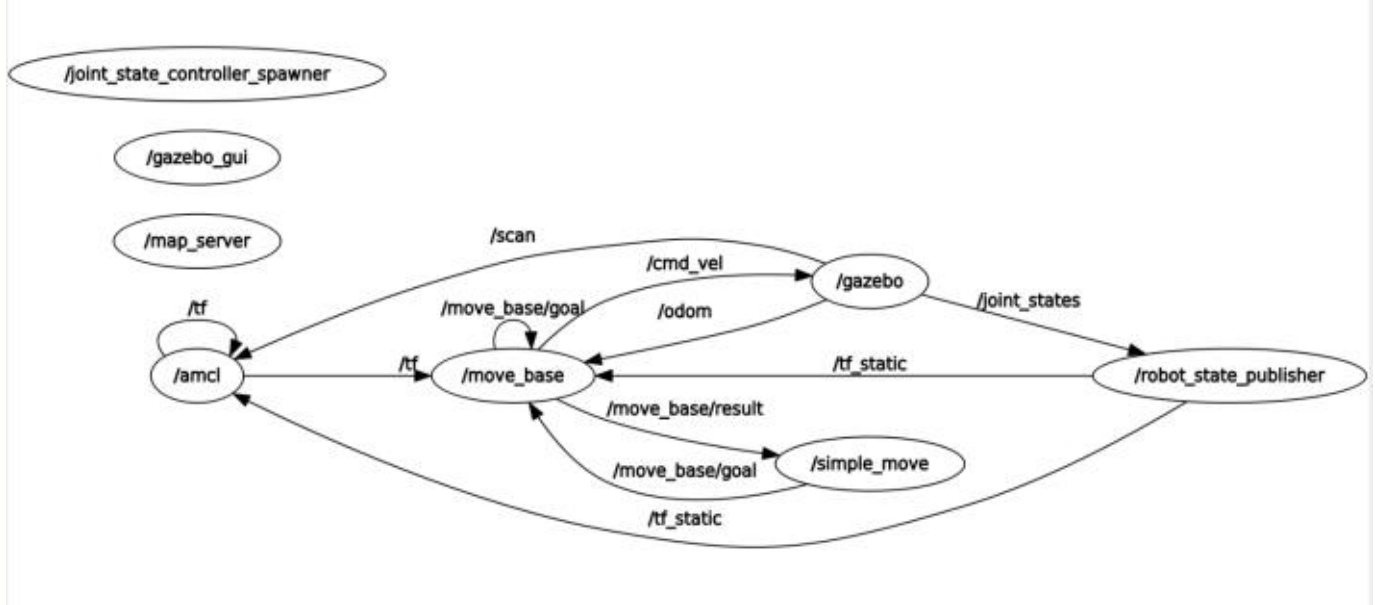

Fig 7. Simple action server implementation

```
def simple_move(x,w):

    rospy.init_node('simple_move')

    #Simple Action Client
    sac = actionlib.SimpleActionClient('move_base', MoveBaseAction )

    #create goal
    goal = MoveBaseGoal()

    #set goal
    rospy.loginfo("Set X = "+x)
    rospy.loginfo("Set W = "+w)

    goal.target_pose.pose.position.x = float(x)
    goal.target_pose.pose.orientation.w = float(w)
    goal.target_pose.header.frame_id = 'base_link'

    #send goal
    sac.send_goal(goal)
```

## A. RESULTS

*Usage:*

***Rosrun minibot3d_sim simple_navi_goals.py 1 1***

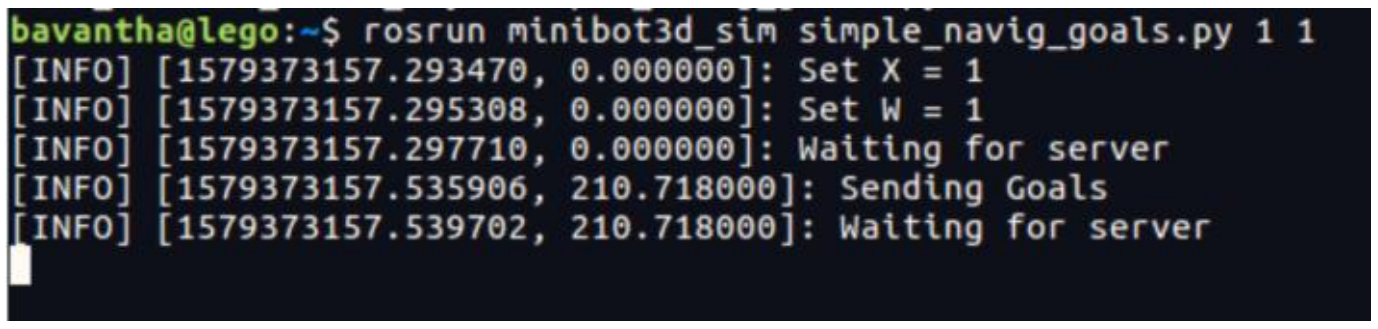

Fig 8. Simple action server terminal output

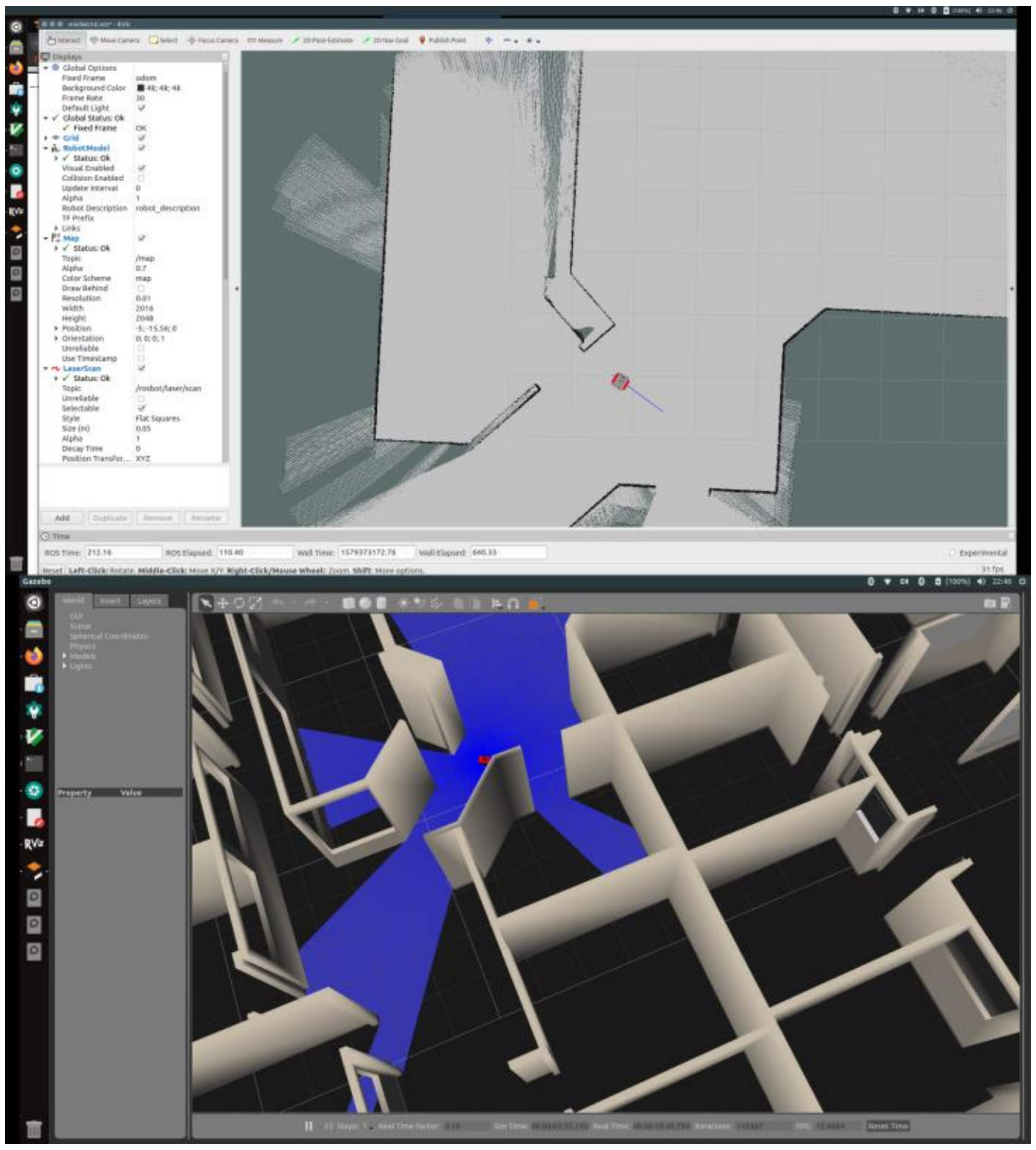

Fig 9. Sample goal using python script

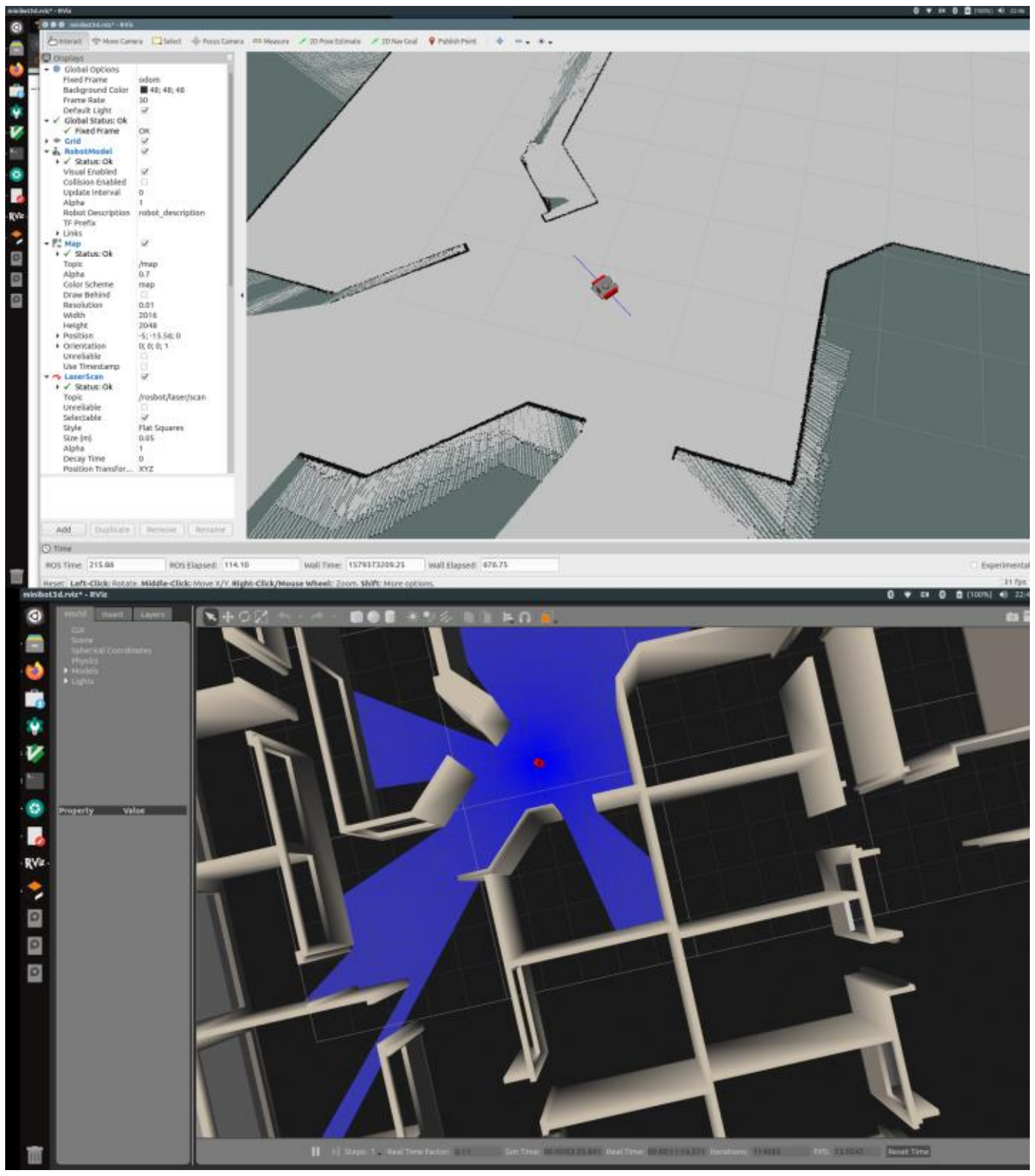

Fig 10. Robot following the given path with the help of Navigation stack

## VI. Parameter tuning: Trajectory Planner tunning

ROS is a global framework for robotics. All the available algorithms are implemented in very reialble manner. Optimization of the parameters should be done before using them in an individual design. Parameter tuning of ROS based implementation is the next major task. Tried three different value set to find out the optimum Trajectory planner parameters. Here the time for the same goal is calculated using custom goal actionlib node.

| Trajectory Planner min_vel_x: | Trajectory Planner max_vel_x: | Performance Sec |
|---|---|---|
| 0.01 | 0.1 | 193 |
| 0.01 | 0.5 | 87 |
| 0.01 | 1 | Never reached and hit a wall |
| 0.1 | 0.1 | 156 |
| 0.1 | 0.5 | 58 |
| 0.1 | 1 | Never reached and hit a wall |

### A. *Min:0.01 and Max : 0.1*

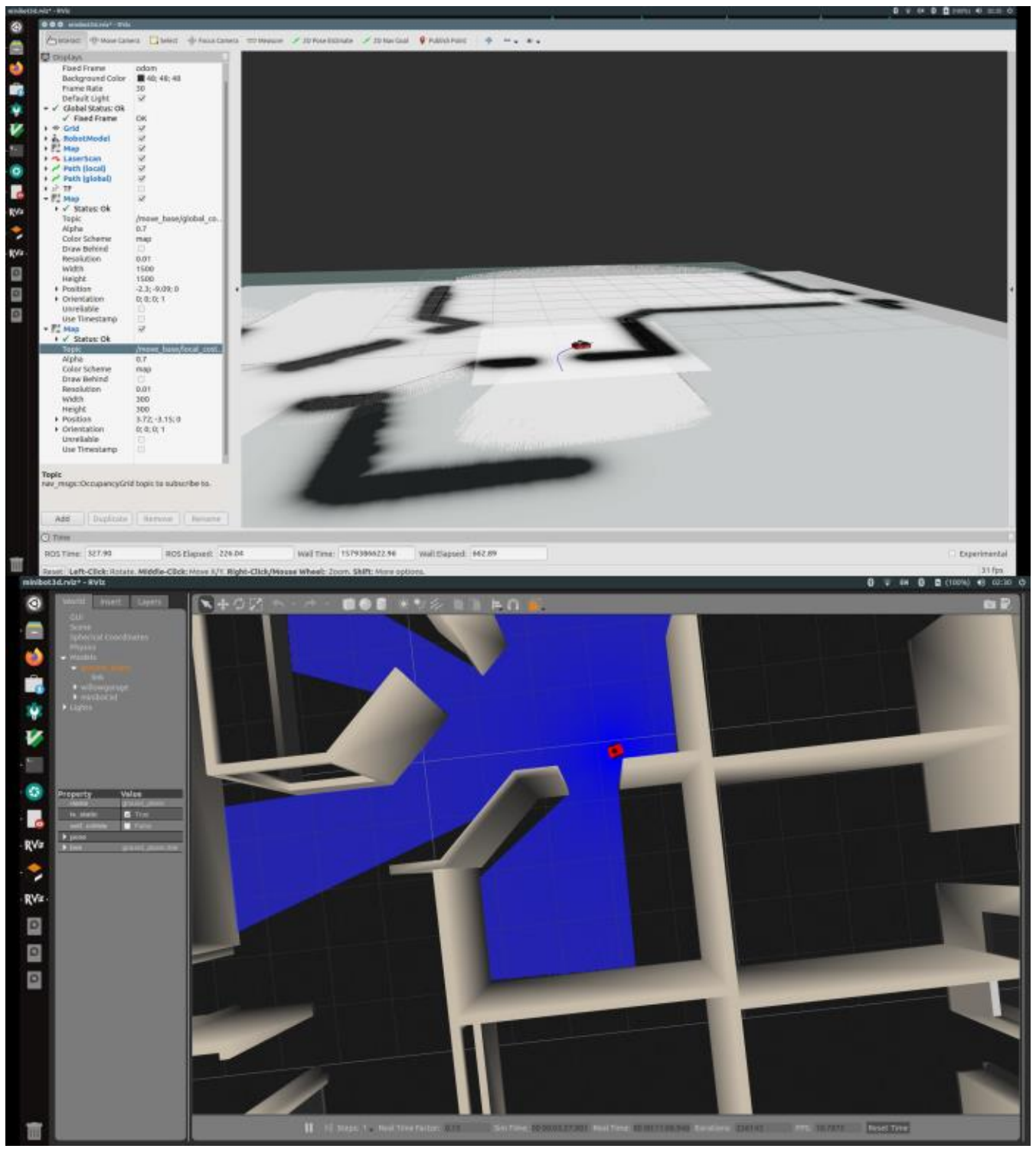

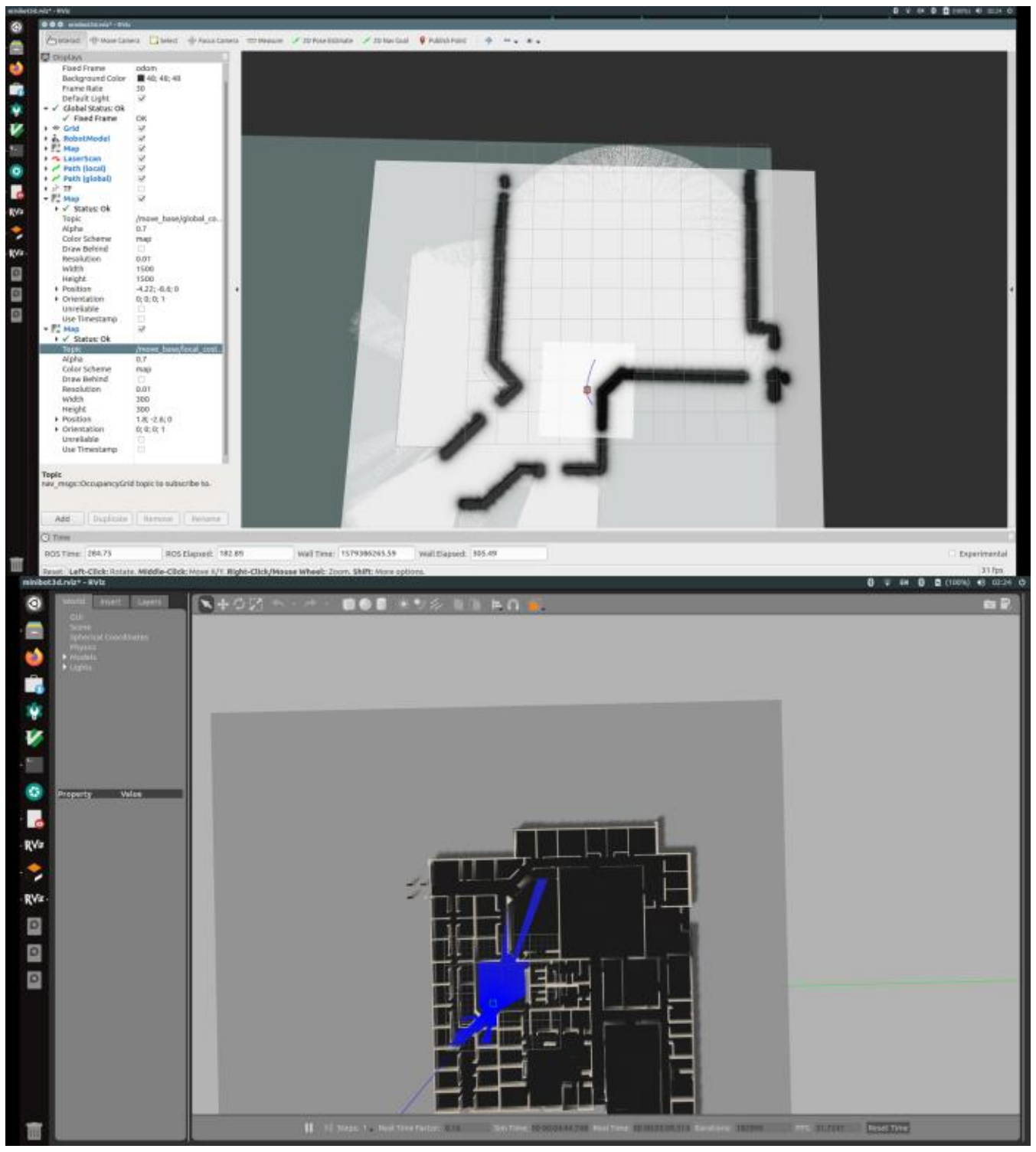

Fig 11. Map with different minimum parameter values

*B. Min:0.1 and Max : 0.5*

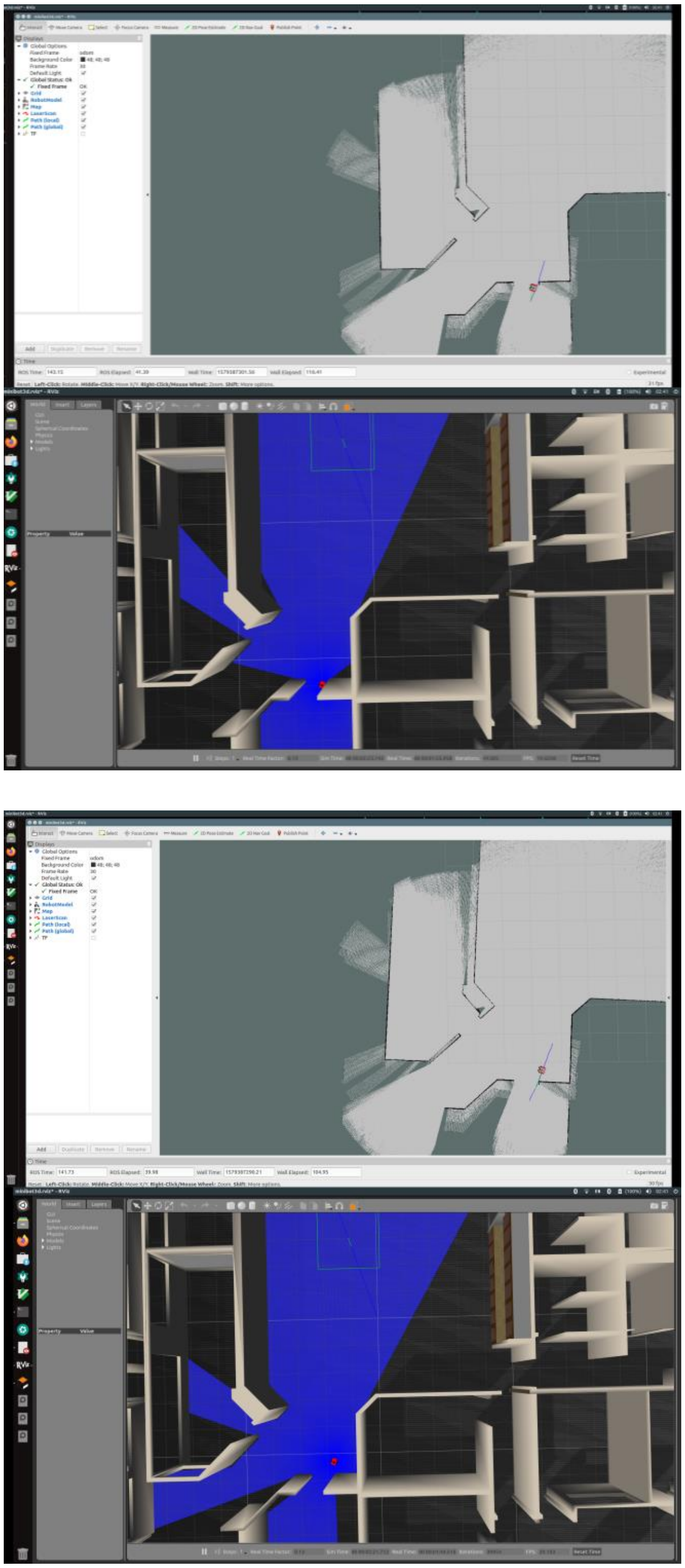

Fig 12. Map with different medium parameter values

*C. Min:0.1 and Max : 0.5*

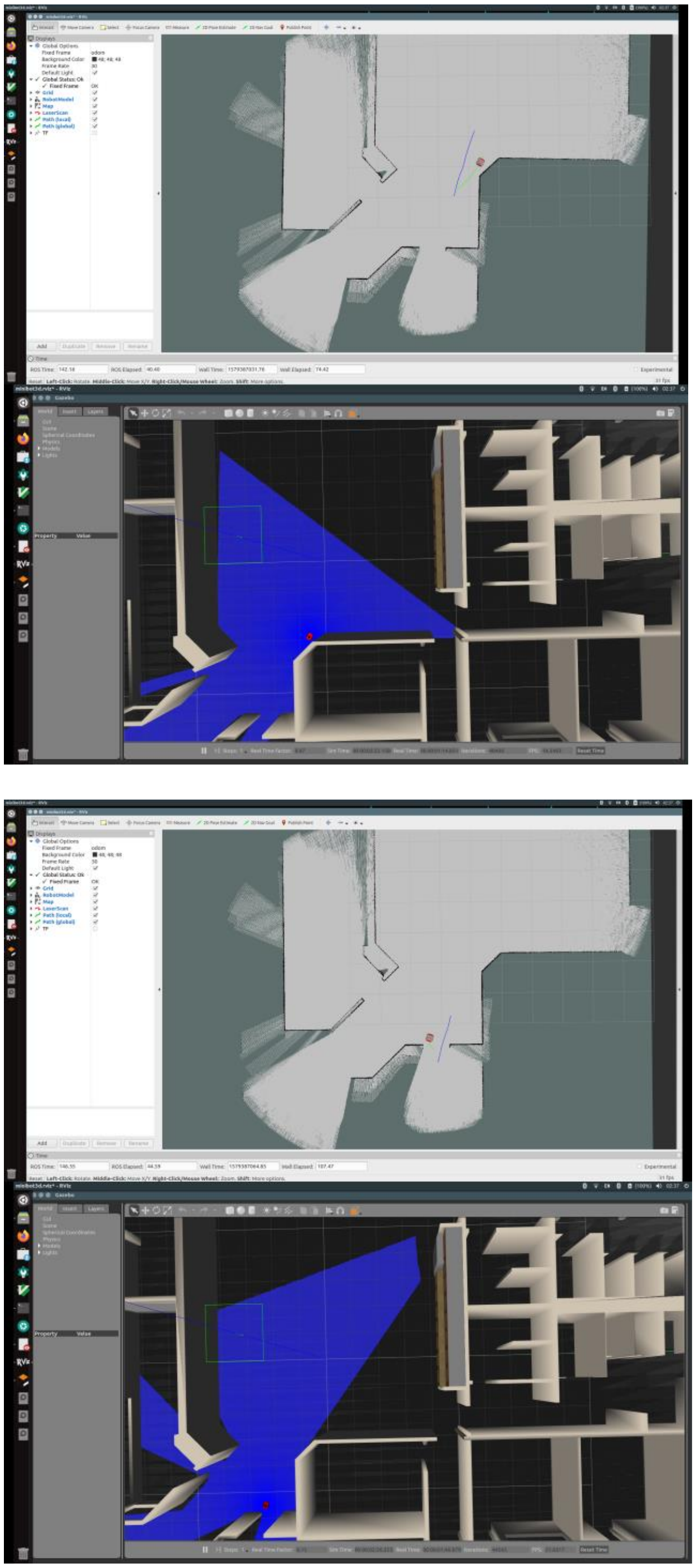

Fig 13. Map with different maximum parameter values

## VII. CONCLUSION

This work presents an experimental evaluation of ROS compatible SLAM algorithm with Gazebo simulation environment. The assessment is set up by executing LIDAR based SLAM algorithm from different parameter values for the trajectory planner. In each experiment, the algorithms according to a established number of criteria enumerated to emphasize practical aspects of the systems on realistic operation scenarios. From the results and discussion this fact is highlighted that Gmapping is an accurate solution for 2D mapping on computationally limited robots with proper tuning of the parameter values for each and every algorithm. That the experiences reported on this work should provide a preliminary guidance for roboticists seeking a SLAM solution for indoor applications.

The ROS framework is a remarkable tool that allows programs to interconnect in very practical ways. A generic Python version of the node able to work for any message type would be a good addition to the ROS topic tools. With an advanced knowledge of ROS and its tools, apparently problematic tasks such as the simultaneous 2D SLAM through a single sensor can be very easy to solve, as long as the problem is addressed from several angles.

Therefore, it can be concluded that the best algorithm to generate maps using the Kinect sensor is Gmapping, because it built approximate maps with an acceptable computational consumption. As future work, one could test the performance of these algorithms using a suitable sensor or inclusive, one could test some algorithms to build maps in 3D.